\documentclass{article}

\usepackage[nonatbib, final]{nips_2017}

\usepackage[utf8]{inputenc} 
\usepackage[T1]{fontenc}    
\usepackage{hyperref}       
\usepackage{url}            
\usepackage{booktabs}       
\usepackage{amsfonts}       
\usepackage{nicefrac}       
\usepackage{microtype}      
\usepackage{graphicx}
\usepackage{amsmath}
\usepackage{subfigure}
\usepackage{pdfpages}
\usepackage{makecell}
\usepackage{todonotes}
\usepackage{color}
\usepackage{caption}
\usepackage{tabularx}
\usepackage{graphicx,subfigure}
\usepackage{cleveref}
\usepackage[numbers]{natbib}
\usepackage{paralist}
\usepackage{relsize}
\usepackage{bbm}
\usepackage{lipsum}  
\usepackage{wrapfig}
\usepackage{varwidth}
\usepackage{blindtext}

\title{Updating the VESICLE-CNN Synapse Detector}

\author{
  Andrew Warrington \\
  Department of Engineering\\
  University of Oxford\\
  England \\
  \texttt{andreww@robots.ox.ac.uk} \\
  \And
  Frank Wood \\
  Department of Engineering\\
  University of Oxford\\
  England \\
  \texttt{fwood@robots.ox.ac.uk} \\ 
}%

\begin{document}

\maketitle


\section*{Introduction}
\label{sec:intro}

Accurate detection of synaptic clefts is paramount to the reconstruction of connectomes~\citep{seung2012connectome}.
One of the most widely cited synapse detection algorithms, VESICLE, is presented by~\citet{roncal2014vesicle}.
VESICLE contains two implementations, a random forest (RF) and a convolutional neural network (CNN) based approach (referred to as V-RF and V-CNN respectively).
V-CNN outperforms the V-RF in terms of classification accuracy, but at the expense of increased computational complexity.
However, the original implementation~\citep{vesiclegit} utilizes a patch-based approach, known to be computationally wasteful due to repeated computations.
Accordingly, this method was suggested to be too computationally intensive for wide-scale application.
Since the VESICLE package is commonly used as a benchmark, we believe the original implementation does not truly reflect the performance that can be obtained by the approach.

Therefore, in this work, we modify the implementation such that it becomes fully convolutional, with no repeated computations, through the use of dilated convolutions, in an architecture we refer to as VESICLE-CNN-2 (V-CNN-2).
These dilated convolutions allow the overall architecture design to remain predominantly unchanged while retaining spatial resolution in the final image.
We then benchmark the performance of our new implementation on modern hardware and quote updated runtime estimates.

Our updated CNN architecture reduces the application speed $600$-fold compared to the runtime on modern hardware, and over a $4500$-fold speedup compared to the originally quoted deployment time.
We release source code and the data used into the public domain, including an updated test-bench for comparing and distributing results.\footnote{Source code and data available at~\url{https://github.com/andrewwarrington/vesicle-cnn-2}.}

\section*{The VESICLE Package}
\label{sec:prior}
By considering synapse detection as the annotation of every voxel in the volume as either synapse or non-synapse, the task can be seen as a semantic segmentation challenge.
This approach is taken by~\citet{roncal2014vesicle} in the VESICLE package.
VESICLE consists of two separate implementations, one utilizing random forests, and the other using CNNs.
 
In this work, we focus on the CNN implementation.
This classifier uses only raw voxel intensities as the input, meaning markedly less preprocessing of data is required compared to the RF implementation that also requires membrane and vesicle labels as inputs.
The architecture for this network is based on the `N3' architecture presented by~\citet{ciresan2012segment}.
This implementation utilizes a patch-based, or sliding window, approach, where a small patch, the size of the field of view of the classifier, is extracted and the CNN applied to that patch.
This approach results in wasted computations as the convolution kernels are repeatedly applied to the same voxels extracted for different patches.
The authors quote this CNN implementation as being $200\times$ slower than the RF implementation, and hence suggest that it is not applicable to large datasets.

\section*{VESICLE-CNN-2}
\label{sec:methods:vesicle}
The first step we take is to modify the original VESICLE-CNN implementation~\citep{roncal2014vesicle} to be a fully convolutional network, referred to as VESICLE-CNN-2.
The patch-based architecture makes use of strided maxpool layers to reduce the spatial dimensions of the patch.
However, applying this maxpooling to a whole image reduces the dimensionality of the image, and hence the output will be of lower resolution than the input.
A one-step method of circumventing this is to use unstrided maxpool operators.
However, this means that the effective field of view is dramatically reduced.
Therefore, we use dilated, or atrous, convolutions.
Atrous convolutions `inflate' the size of the mask by inserting fixed zeros at regular intervals.
For instance, dilation of a three by three mask, with a dilation rate of one yields a five by five mask, with two rows and columns of zeros (resembling a noughts-and-crosses board).
This allows us to capture the spatial extent similar to the original implementation, while retaining a{%
\parfillskip=0pt
\parskip=0pt
\par}
\begin{wrapfigure}[38]{R}{0.37\textwidth}

 \begin{minipage}[h]{\linewidth}
 	\centering
	\includegraphics[width=\textwidth]{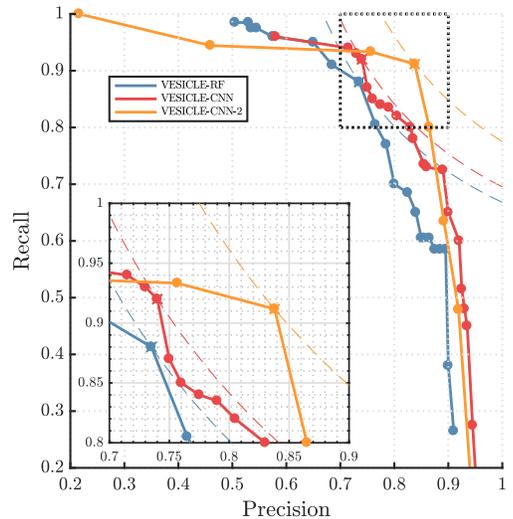}
	\caption{Precision-recall (P-R) characteristics for whole-synapse detection when applied to the validation set. Dashed lines represent constant F1 contours, set to the maximum F1 score achieved by each classifier, while dots represent observed operating points. We were unable to replicate the results of~\citet{roncal2014vesicle} ourselves and hence quote results read from the original paper.}
	\label{fig:obj_pr}
  \end{minipage}

\vspace*{0.5cm}

  \begin{minipage}[h]{\linewidth}
    \centering
	\begin{tabular}{@{}lllllll@{}}
		\toprule
		Architecture 							& \multicolumn{2}{c}{Execution time} 	& Test F1 										\\
 														& 	\multicolumn{2}{c}{(minutes)}			& 														\\
 		\cmidrule{2-3}
 														& Training					& Deploy					& \\
		\midrule
		V-RF				 							& \textbf{6.15}	 	&  12.9					& 0.801												\\
		V-CNN										& 512 						&  290 					& 0.820											\\	
		V-CNN-2				  					& 565		          		&  \textbf{0.493}	& \textbf{0.869}								\\
		\bottomrule			
	\end{tabular}							
	\captionof{table}{Comparison of performance and run-time of the different algorithms used. Bold text indicates the best performing algorithm in each category. Time requirements evaluated on a single multicore desktop machine, equipped with $6\times $ Intel Core i7-5930K, 3.7 GHz ($12$ logical cores), $64$GB RAM and a new generation Nvidia Pascal Titan Xp GPU with $12$GB onboard memory. All times are quoted for execution on a single thread and hence are maximal runtimes.}		
	\label{tab:results}
 \end{minipage}%
 \end{wrapfigure}
\noindent fully convolutional structure, but without exposing us to overfitting if we were to simply use larger, undilated convolutional kernels.

Training and evaluation is conducted using the same methodology as was used by~\citet{roncal2014vesicle} and is as follows:
Training, validation and test volumes are non-overlapping volumes drawn from the~\citet{kasthuri2015saturated} dataset, imaged at $3$nm$\times3$nm$\times30$nm.
Images are $1024\times1024$ voxels in the imaging plane.
Training, validation and test volumes are composed of $75$, $25$ and $100$ images respectively.
Hyperparameter optimization was performed on the validation set.
These hyperparameter sweeps are shown in Figure~\ref{fig:obj_pr}.
This optimization is required by all scripts and hence we do not include the time required in the training time for each algorithm.

\section*{Results}
\label{sec:results}

We now compare the performance of our VESICLE-2 to the original VESICLE implementation. 
We were unable to replicate the results for VESICLE and hence the results we quote for VESICLE are read from the original text, with test F1 score (Table~\ref{tab:results}) being quoted as the maximum value from this graph (an upper bound as generalization means this score can only reasonably decrease when moving away from the validation set).

Precision-recall curves for whole-synapse detection are shown in Figure~\ref{fig:obj_pr} and shows our network performs at least as well as the original implementation, if not better, achieving a higher operating point in terms of F1 score.
This magnitude of this improvement is shown in Table~\ref{tab:results}. 
The source of this improvement may be due to using an improved optimizer or data being drawn from a different subvolume of the Kasthuri dataset~\cite{kasthuri2015saturated}.
The slight alteration in architecture specifics (due to implementation requirements) is unlikely to have induced this change, and hence we continue to describe this as an `update' as opposed to a new architecture.
However, as  desired, the train and deployment times, as also shown in Figure~\ref{tab:results}, are dramatically lower than the original implementation, by a factor of $600$.
This improvement makes the application of the VESICLE algorithm to large datasets viable, and hence shifts the benchmark for any new algorithm that aims to supersede this work.

\section*{Conclusion}

In this short paper we have presented an updated version of the synapse detection algorithm presented by~\citet{roncal2014vesicle}.
We have shown that by using dilated convolutions it is possible to create a fully convolutional approximation to the original, patch-based CNN implementation, bringing vast reductions in application time.
We have also made source code available for our implementation, as well as updated test beds for comparison of the performance of each algorithm.

\begin{small}
\bibliographystyle{unsrtnat}
\bibliography{synapse_detection}{}

\begin{thebibliography}{5}
\providecommand{\natexlab}[1]{#1}
\providecommand{\url}[1]{\texttt{#1}}
\expandafter\ifx\csname urlstyle\endcsname\relax
  \providecommand{\doi}[1]{doi: #1}\else
  \providecommand{\doi}{doi: \begingroup \urlstyle{rm}\Url}\fi

\bibitem[Seung(2012)]{seung2012connectome}
S.~Seung.
\newblock \emph{Connectome: How the Brain's Wiring Makes Us who We are}.
\newblock A Mariner Book. Houghton Mifflin Harcourt, 2012.
\newblock ISBN 9780547508184.

\bibitem[Roncal et~al.(2014)Roncal, Pekala, Kaynig-Fittkau, Kleissas,
  Vogelstein, Pfister, Burns, Vogelstein, Chevillet, and
  Hager]{roncal2014vesicle}
W.~Roncal, M.~Pekala, V.~Kaynig-Fittkau, D.~Kleissas, J.~Vogelstein,
  H.~Pfister, R.~Burns, R.~Vogelstein, M.~Chevillet, and G.~Hager.
\newblock Vesicle: Volumetric evaluation of synaptic interfaces using computer
  vision at large scale.
\newblock \emph{Proceedings of the British Machine Vision Conference (BMVC),
  pages 81.1-81.13. BMVA Press, September 2015}, 2014.

\bibitem[Roncal(2015)]{vesiclegit}
W.~Roncal.
\newblock Vesicle synapse detection repository.
\newblock \url{https://github.com/neurodata/vesicle}, 2015.

\bibitem[Ciresan et~al.(2012)Ciresan, Giusti, Gambardella, and
  Schmidhuber]{ciresan2012segment}
D.~Ciresan, A.~Giusti, L.~Gambardella, and J.~Schmidhuber.
\newblock Deep neural networks segment neuronal membranes in electron
  microscopy images.
\newblock In F.~Pereira, C.~J.~C. Burges, L.~Bottou, and K.~Q. Weinberger,
  editors, \emph{Advances in Neural Information Processing Systems 25}, pages
  2843--2851. Curran Associates, Inc., 2012.

\bibitem[Kasthuri et~al.(2015)Kasthuri, Hayworth, Berger, Schalek, Conchello,
  Knowles-Barley, Lee, Vazquez-Reina, Kaynig, Jones, Roberts, Morgan, Tapia,
  Seung, Roncal, Vogelstein, Burns, Sussman, Priebe, Pfister, and
  Lichtman]{kasthuri2015saturated}
N.~Kasthuri, K.~Hayworth, D.~Berger, R.~Schalek, J.~Conchello,
  S.~Knowles-Barley, D.~Lee, A.~Vazquez-Reina, V.~Kaynig, T.~Jones, M.~Roberts,
  J.~Morgan, J.~Tapia, H.~Seung, W.~Roncal, J.~Vogelstein, R.~Burns,
  D.~Sussman, C.~Priebe, H.~Pfister, and J.~Lichtman.
\newblock {{S}aturated {R}econstruction of a {V}olume of {N}eocortex}.
\newblock \emph{Cell}, 162\penalty0 (3):\penalty0 648--661, Jul 2015.

\end{thebibliography}
\end{small}

\end{document}